\documentclass[letterpaper, 10 pt, conference]{ieeeconf}

\IEEEoverridecommandlockouts   
\usepackage{dcolumn} 
\usepackage[table]{xcolor} 
\usepackage{makecell}
\usepackage{hhline}

\usepackage{amsmath}
\usepackage{adjustbox}
\usepackage[ruled,vlined]{algorithm2e}
\usepackage{arydshln}
\usepackage{array}
\usepackage{balance}
\usepackage{booktabs}
\usepackage{colortbl}
\usepackage{comment}
\let\labelindent\relax
\usepackage{enumitem}
\usepackage{etoolbox}
\usepackage{graphicx}
\usepackage{hyperref}
\usepackage{listings}
\usepackage{makecell}
\usepackage{multicol}
\usepackage{multirow}
\usepackage{nameref}
\usepackage{soul}
\usepackage{capt-of}
\usepackage{tabularx}
\usepackage{textcomp}
\usepackage{threeparttable} 
\usepackage{tikz}
\usepackage{url}

\definecolor{orange0}{HTML}{FFF0E5} 
\definecolor{orange1}{HTML}{FFE0C7} 
\definecolor{orange2}{HTML}{FFCC99} 
\definecolor{orange3}{HTML}{FFA630} 
\definecolor{orange4}{HTML}{F17720} 

\definecolor{blue0}{HTML}{E5F8FF}   
\definecolor{blue1}{HTML}{C0EAF9}   
\definecolor{blue2}{HTML}{99D7F4}   
\definecolor{blue3}{HTML}{00A7E1}   
\definecolor{blue4}{HTML}{0474BA}   

\definecolor{gray1}{HTML}{EBEBEB}
\definecolor{gray0}{HTML}{F7F7F7}   
\definecolor{gray2}{HTML}{D5D5D5}   

\def\eg{\emph{e.g., }} 

\def\ie{\emph{i.e., }} 

\newcommand{\insertfig}{\vspace{1em}\includegraphics[width=\linewidth]{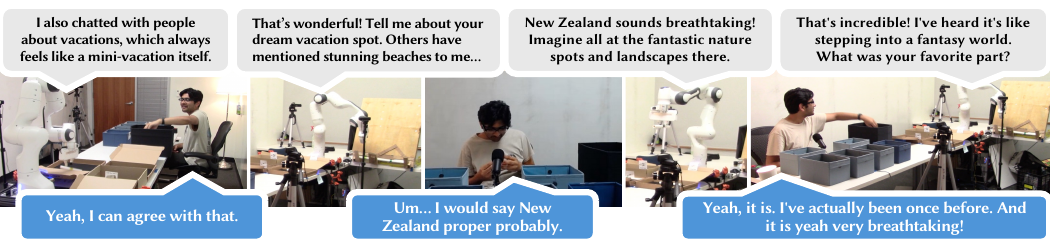}\captionof{figure}{We investigate users' interactions with an autonomous manipulator that engages in small talk with varying levels of self-disclosure during a collaborative task. The example here shows a participant in the high-disclosure condition [H245]. 
}
\label{fig:teaser}}
\makeatletter
\apptocmd{\@maketitle}{\centering\setcounter{figure}{0}\insertfig}{}{}
\makeatother

\definecolor{dihl}{RGB}{159,191,229}
\newcommand{\designimp}[2]{%
  \par\smallskip
  \noindent\colorbox{dihl}{\bfseries\color{black} #1}%
  \par\nobreak\vspace{1pt}%
  \noindent\textit{#2}\par\smallskip}
  
\title{\LARGE \bf
From Small Talk to Rapport: \\Exploring Robot Self-Disclosure in Collaborative Tasks
}
\author{Kaitlynn Taylor Pineda$^{1}$$^{2}$, 
Anvii Mishra$^{1}$,
Brian Chien$^{1}$, 
Angela Guo$^{1}$, 
Toluwani Williams$^{1}$, \\
Ziang Xiao$^{1}$, 
and Chien-Ming Huang$^{1}$
\thanks{$^{1}$Johns Hopkins University, Department of Computer Science, Baltimore, MD, USA} 
\thanks{$^{2}${\tt\small kpineda3@jhu.edu}}
}%

\begin{document}
\maketitle
\begin{abstract}
People naturally chat while collaborating and share personal information (\ie self-disclose) to build rapport and maintain social connections. 
As robots are increasingly developed to work with people, the effective use of these social behaviors to enhance engagement and support teamwork becomes ever more important. 
While prior work has shown that robot-initiated small talk can benefit human–robot collaboration, less is known about how best to design such small talk. In this work, we explore how self-disclosure may be designed to support small talk within a human–robot team---especially when the robot is an industrial manipulator that lacks anthropomorphic cues and performs physical work.
We first developed an LLM-driven manipulator capable of partaking in small talk, adopting either a \textit{low-disclosure} or \textit{high-disclosure} strategy.
We then conducted a user study ($N = 50$) to investigate how self-disclosure in small talk influences human–robot dynamics. Unexpectedly, participants disclosed more in the low-disclosure condition and reported stronger teaming and coordination than those in the high-disclosure condition. This effect was more pronounced among users with prior experience teaming with robots.
These results suggest that increasing robot self-disclosure does not necessarily foster rapport, social connection, or reciprocal disclosure; other factors, such as prior HRI experience, should be considered.
\end{abstract}

\maketitle


\section{Introduction}
Non-anthropomorphic robots, such as industrial manipulators, are among the most commonly deployed robot platforms in human–robot collaboration settings like manufacturing and logistics \cite{ghodsian2023mobile}. 
These robots are typically designed to assist with physically demanding or precision tasks alongside people. 
Recent work \cite{pineda2025see} has shown that adding social communication, like small talk, that accompanies physical collaboration can improve people’s engagement with the robot and their rapport with it compared to no small talk. 
However, it remains unclear how best to design such small talk to enhance human–robot team dynamics: what content and style are most effective for engaging in small talk?

Drawing on human interpersonal communication, 
self-disclosure---defined as the voluntary sharing of personal thoughts, feelings, or information---plays a key role in building rapport and trust \cite{greene2006self,altman1973social}. 
Prior HRI studies have examined disclosure behaviors and reported that they can elicit reciprocal disclosure and strengthen perceived social bonds \cite{martelaro2016tellmemore,eyssel2017disclosure}. 
Those studies, however, often involved humanoid or social robots, and it is unknown whether the same effects generalize to non-anthropomorphic robots whose physical form is clearly functional rather than social.

In this work, we investigate whether and how disclosure-oriented small talk from a non-anthropomorphic manipulator shapes human partners’ responses during collaboration, particularly their willingness to self-disclose and their perceptions of the robot. 
We developed an autonomous manipulator capable of LLM-driven small talk and evaluated it with 50 participants in a co-working scenario. 
Participants interacted with either a low-disclosure or high-disclosure robot, allowing us to examine how the robot’s disclosure level affects user self-disclosure, rapport, and conversation dynamics.
Our results provide an empirical characterization of how differences in robot self-disclosure influence user self-disclosure and perceptions of non-anthropomorphic robots during physical tasks, and offer design implications.

\section{Background and Related Works}
\label{section:background}


\subsection{Self-Disclosure Through Small Talk}
Social penetration theory suggests that self-disclosure gradually increases over the course of a relationship and serves as a mechanism for building closeness and trust \cite{altman1973social}.
Light, everyday conversations such as small talk \cite{Laver1981LinguisticRA, sprecher2018self} often provide the setting in which such disclosures emerge, beginning with superficial information and deepening over time. 
Although commonly associated with established relationships, disclosure also supports early relationship formation, with small talk providing a low-stakes context for incremental and reciprocal information sharing that gradually builds familiarity and trust \cite{sprecher2018self}. 
Such sharing promotes the development of close bonds between people \cite{altman1973social} and provides a means for individuals to learn more about each other \cite{sprecher2018self}.
Conversational partners also commonly use elicitation strategies (\eg questions) to 
prompt their interlocutor to expand or add detail within a single turn \cite{Lerner2004Prompting}.
Elicitation strategies provide a way to not only extend conversational turns but also 
extract more information from one's partner, serving as an important mechanism in encouraging self-disclosure in conversations and increasing intimacy.
\subsection{Self-Disclosure in Dialogue Systems}
Self-disclosure strategies have been implemented in conversational dialogue systems to enhance user engagement and response quality. 
For example, emotional disclosures have been shown to encourage reciprocal sharing, increase enjoyment, and improve perceptions of dialogue agents 
\cite{liang2024Dialoging, Mou2024Liking, Sauppé_Mutlu_2015}.
Furthermore, prior work \cite{augustine2024motives} found that participants were more likely to share highly intimate or negative disclosures with robots than with humans, possibly due to people viewing robots as non-judgmental companions. 
%
Together, these studies suggest that robots designed to elicit self-disclosure can foster more positive interactions \cite{liang2024Dialoging}. 
However, prior work has mostly focused on examining humanoid, social, or agent-like systems, where disclosure may more readily align with users' expectations of the robot as a social partner \cite{chen2024effects,martelaro2016tellmemore}.

It remains an open question how these findings extend to non-anthropomorphic robotic manipulators. 
Prior work has shown that robot-initiated small talk can benefit human–robot collaboration even with non-anthropomorphic manipulators \cite{pineda2025see},
yet it is unclear how best to design such small talk—particularly how self-disclosure should be used within it.
In particular, it is unclear how people would respond to, and perceive, non-anthropomorphic manipulators---whose embodiment is more strongly associated with function than sociality---when they engage in disclosure.
Indeed, in collaborative settings, people may view robots as functional tools or task-oriented teammates \cite{mingyue2017human,ososky2012importance}.
Furthermore, robot disclosures are authored design choices rather than expressions grounded in genuine lived experience \cite{dietrich2023disclose,Mou2024Liking,neerincx2022disclosure}.
Therefore, whether and how disclosure can be used effectively to promote positive human–robot interaction in collaborative settings, where non-anthropomorphic robots remain commonplace, requires further study.
In this work, we examine how disclosure design choices in small talk shape perceptions of a robotic manipulator, exploring how minimal persona cues conveyed through disclosure may influence interaction outcomes.

\section{Robot System}
\begin{figure}[h]
    \centering
\includegraphics[width=0.45\textwidth]{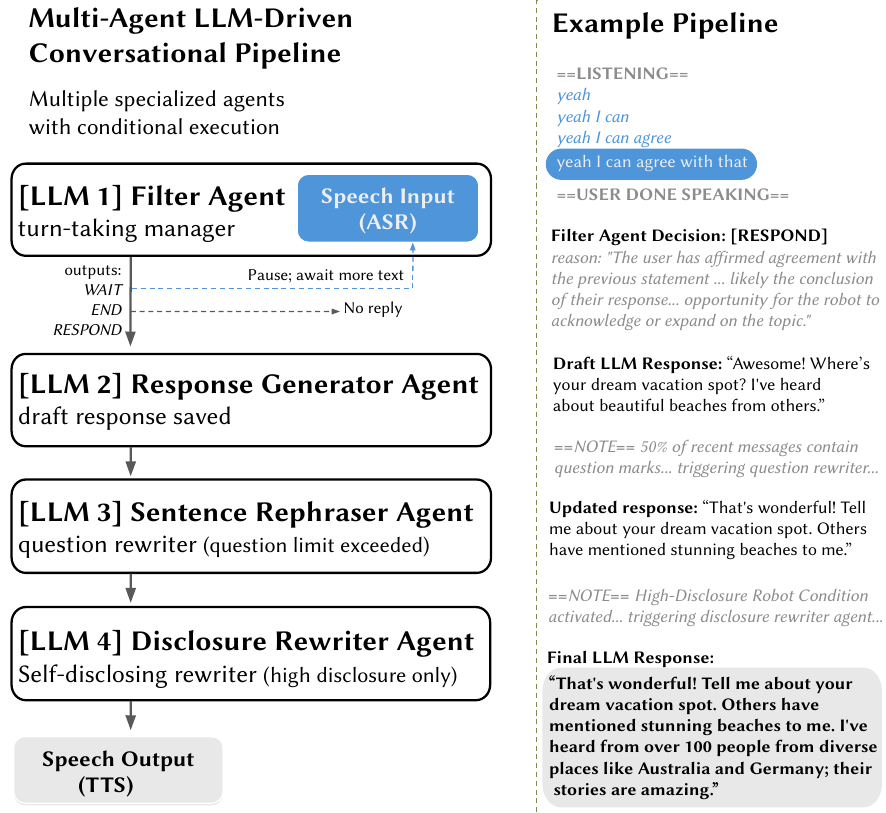}
    \caption{ Left: Conversational workflow of the system using multiple helper LLM agents. 
    Right: Detailed pipeline example for a HD participant [H245].}
    \label{fig:system}
\end{figure}

The autonomous robot system consists of a speech listener (ASR), a GPT-4o conversation generation pipeline, and a 7-DOF Franka Emika Panda manipulator. 
The conversational
pipeline (See Fig. \ref{fig:system}) handled spoken interaction, while the robot’s physical
behavior was executed separately through pre-programmed motion trajectories.
During each sub-task cycle, the robot
picked up a PVC pipe, 
inspected it, placed it into the
appropriate sorting location (yellow, green, or faulty), and
returned to a home pose before beginning the next cycle. 
These task motions followed the same fixed sequence for each
participant.
Additional details on the conversational pipeline are described below. 

\subsection{Speech Listener: Automatic Speech Recognition (ASR)}
An ASR stream converts user speech to text using the Google Cloud Speech-to-Text API. 
Audio is processed in 2.6-sec. chunks at a 16 kHz sampling rate
with silence marking response completion; this set up is similar to our prior work \cite{pineda2025see}.
The ASR pauses while the LLM processes user input and the robot speaks. 
\subsection{LLM-based Conversation Generation} 
Transcribed text is sent into a conversational workflow of multiple LLM-powered components (Fig. \ref{fig:system}); for complete LLM agent prompts, see Appendix E\footnote[3]{https://tinyurl.com/ro-man-appendix}.


\begin{enumerate}[leftmargin=*]
    \item \textbf{Filter Agent.} 
    Manages appropriate turn-taking by classifying the user's transcribed speech as 
    \textit{Wait} (response still in progress), 
    \textit{Respond} (response complete, pass downstream), or 
    \textit{End} (no reply needed). 
    This ensures the workflow advances only when user input is ready.
    \item \textbf{Response Generator Agent.} Produces a candidate response tailored to the user’s input, configuring the robot’s conversational style (\eg extroverted tone, concise length, alignment with participant affect)
    to serve as a base utterance for downstream refinement. 
    \item \textbf{Sentence Rephraser Agent.} Promotes balanced conversational flow by limiting excessive robot questioning;
    monitors robot question frequency 
    and, if it exceeds 50\%, 
    rewrites the candidate response into a non-question.
    \item \textbf{Disclosure Rewriter Agent.} In the high-disclosure condition, this agent enables explicit robot self-disclosure by rewriting the candidate response to include a contextualized reference to a preset list of fictional robot experiences, represented as summaries of prior conversations. 
    This agent is not present in the low-disclosure condition. 
\end{enumerate}

\section{Evaluation}
\subsection{Study Task}
We designed the experimental task as cooperative 
rather than strictly collaborative.
Framed within a single simulated factory, 
the human and robot shared a quality-control goal but pursued it through a division of labor: 
the (pre-programmed) robot sorted PVC pipes by color
while the participant inspected and sorted PVC fittings, rather than synchronous joint manipulation \cite{kolbeinsson2019foundation}.
As it sorted, the robot also initiated chit-chat with the user.
We chose this loosely-coupled design deliberately, as it reduces physical coordination demands and isolates the robot's conversational behavior, 
providing a controlled setting for examining how robot dialogue behaviors,
such as self-disclosure, 
shape user experience during 
cooperative work. 
In contrast to our prior, more tightly-coupled collaborative work \cite{pineda2025see}, 
this task foregrounds social over physical coordination. 
Participants set aside `faulty' (black dot marking) PVC fittings while sorting by shape;
on average, the task took approximately 
17 minutes.

\subsection{Study Conditions}
Participants were randomly assigned to one of two conditions (described below), and the main task officially began after users pressed a green start button. For each participant, the robot initiated small talk speech about four seconds after the green button press. The first spoken phrase by the robot was always: \textit{``Hey there, how is your day going so far?"}. 
Additional details on the differences in robot persona across conditions can be seen in Appendix E\footnotemark[3]. 
\begin{itemize}[leftmargin=*]
    \item \textbf{Low Disclosure (LD).} 
    Robot was given a basic persona with minimal personal information, and instructed
    to share minimal personal details about itself with the user.
    \item \textbf{High Disclosure (HD).}
    Robot was given an expanded persona containing additional personal details.
    Prior work in HRI highlights the need to carefully design robot persona and autonomy in ways that remain coherent with the system’s role and capabilities \cite{neerincx2019socio,clabaugh2019escaping}. Consequently, we operationalize high disclosure using plausible, contextually grounded details; `fictional' lived experiences or `fictional' summaries of prior  conversations, rather than more strongly anthropomorphic or physically incongruent attributes. Fig. \ref{fig:teaser} illustrates an example.

\end{itemize}
\subsection{Study Procedure}
Participants completed consent forms and pre-study surveys: 
demographics (Sec. \ref{section:participants}) and 
self-disclosure propensity \cite{IPIP_RD3}. 
The experimenter provided verbal and written task instructions before showing the user a brief demo of the robot's speech capabilities and range of motion.
Participants sorted their designated box of PVC fittings by shape and `faultiness' in the main task, during which 
the experimenter left the room.
Users indicated the official start and stop of their task via buttons on the table. 
When done, 
users completed a post-study survey and semi-structured interview.

\subsection{Participants}
\label{section:participants}
We recruited 50 participants, with 25 (14 male, 11 female) in each condition; overall ave. age of 23.66 (SD = 8.29). All were compensated at \$15.00 USD per hour. 
On 1--7 scales, participants reported mean scores of 
4.41 ($SD=0.78$) for personality (self-disclosure),
5.78 ($SD=1.20$) for experience with technology,  
4.6 ($SD=1.64$) for experience with voice assistants, 
3.66 ($SD=1.73$) for experience with robots, 
and 
2.36 ($SD=1.52$) for experience with teaming with robots.
There were no significant differences between conditions on any of these scales. 
This study received university IRB approval.

\section{Measures}
\label{section:measures}
Conversational dynamics may extend beyond what any single metric can capture. 
For instance, when establishing effective small talk,
prior work indicates that concise responses, maintaining a positive tone, covering a broad range of topics, and yet remaining contextually relevant are key elements \cite{ramnauth2024more}. 
We therefore designed measures to capture not only the amount of user disclosure, but also how that disclosure emerged during interaction.
Specifically, we examined whether user disclosures were explicitly elicited by the robot or arose spontaneously without direct prompting, a distinction grounded in conversational elicitation theory and prior work on reciprocal disclosure in dialogue systems \cite{pillet2018begin,ravichander2018empirical}. 
This provides a more nuanced view of reciprocity by showing whether robot disclosure strategies influence not only how much users disclose, but also the interactional conditions under which disclosure occurs.
\subsection{Manipulation Check} 
\label{manipulation-check}
The metrics below help confirm the robot's speech patterns aligned with the intended low/high disclosure personas. 
\begin{itemize}[leftmargin=*]
    \item \textbf{Robot Word Count}. The total number of words spoken by the robot during main task interaction.
    \item \textbf{Number of Robot Turns}. The total number of conversational turns by the robot during main interaction.
    \item \textbf{Robot Self-Disclosure Score.} 
    All spoken robot utterances were coded into one of three self-disclosure levels: 
    \textit{0-no disclosure},
    \textit{1-low disclosure}, or 
    \textit{2-high disclosure},  
    with higher levels reflecting more personal and elaborated disclosure.
    For the main task interaction, we calculated the proportion of robot utterances in each disclosure category. 
    Coding details can be found in Appendix A\footnotemark[3].
\end{itemize}

\subsection{Conversational Measures}
\label{conversational-measures}
We measured user disclosure levels, 
whether it was robot-elicited or spontaneous,
and broader interaction patterns.
\begin{itemize}[leftmargin=*]
\item \textbf{User Self-Disclosure Score.}
All spoken user utterances were coded into the same three self-disclosure levels: 
\textit{0-no disclosure},
\textit{1-low disclosure}, or 
\textit{2-high disclosure}, 
with higher levels reflecting more personal and elaborated disclosure.
For the main task interaction, we calculated the proportion of user utterances in each disclosure category.
\item \textbf{(Un)Prompted Responses (\%).} 
User and robot utterances were labeled as an independent statement (unprompted) or a direct (prompted) response to a preceding statement. We recorded the totals of each over the total turns per speaker. 
Further details distinguishing between (un)prompted utterances can be found in Appendix A\footnotemark[3].
\item \textbf{User Word Count.} 
The total number of words spoken by the user during the main interaction.
\item \textbf{Number of User Turns.} 
The total number of conversational turns by the user during the main interaction.
\item \textbf{Utterance Type (\%).} 
All spoken utterances were classified as either \textit{questions}, \textit{imperatives}, or \textit{non-questions (comments/statements)}; the total of each type during the task was found for the user and robot, and calculated as a percentage over total turns for each speaker.
Further details defining utterance types can be found in Appendix B\footnotemark[3].

%
\end{itemize}


\subsection{Subjective Measures (User Perceptions)} 
\label{subjective-measures}
%
Unless stated otherwise, all constructs below were formed via scale construction and used a 1--7 scale for responses:
See the Appendix\footnotemark[3] for individual items for each scale construct.  
 \begin{itemize}[leftmargin=*]
  \item \textbf{Disclosure} (5 items; Cronbach's $\alpha = 0.80$).
  Users' perceived self-disclosure to the robot over the main task.
  \item \textbf{Teaming} (5 items; Cronbach's $\alpha = 0.85$).
  Measures users' perception of teaming with the robot over the main task.
  \item \textbf{Boredom} (2 items;  Cronbach's $\alpha = 0.76$).
   Measures users' perceived boredom over the main task.
   \item \textbf{Rapport.} We used the Connection-Coordination Rapport (CCR) scale to evaluate users' perceived rapport with the robot \cite{lin2025connection}; this measure used a 1--5 scale: \textit{Rapport-Connection} (12 items; 
        Cronbach's 
        $\alpha = 0.90$) and \textit{Rapport-Coordination} (6 items; 
        Cronbach's 
        $\alpha = 0.84$).
    
\end{itemize}

\section{Data Analysis}
\label{data-analysis}
Study transcriptions from system logs were manually verified against video recordings.
Each utterance was labeled for
\textit{utterance type} 
and
\textit{self-disclosure level}
based on the definitions in Sec. \ref{section:measures} and Appendices A \& B\footnotemark[3].
To establish reliability, two human raters independently labeled an initial 10\% data. 
Agreement improved after discussion and refinement of disagreements (see Appendix C\footnotemark[3]),
reaching Cohen's 
$\kappa = 0.909$ for \textit{utterance type} labels, 
$\kappa = 0.900$ for \textit{self-disclosure level} labels,
and an overall $\kappa = 0.833$.
ChatGPT-5 served as a third rater, using the same labeling instructions as the authors, yielding three rater Fleiss' Kappa values of
$\kappa = 0.817$ for \textit{utterance type} labels
and 
$\kappa = .726$ for \textit{self-disclosure} labels.

Turn-level \textit{user self-disclosure} was 
treated as an ordinal three-level outcome (no disclosure, low disclosure, high disclosure), 
with higher levels reflecting more personal and elaborated disclosure. 
We analyzed this outcome using a cumulative link mixed-effects model (CLMM) with participant-level random intercepts to account for repeated conversational turns within participants.
Fixed effects included \textit{experimental condition} (LD/HD), \textit{user (un)prompted responses}, and their interaction.
For continuous post-study survey scales,
we fit linear regression models 
with
\textit{experimental condition} 
and 
pre-study participant characteristics as predictors. 
Candidate covariates included 
\textit{age, gender, prior experience with technology, prior experience with robots,} 
and \textit{prior experience teaming with robots}. 
Continuous covariates were z-scored before analysis.
Final models were selected using AIC-based stepwise regression following established analytic approaches \cite{huang2014multivariate}.
Including participant covariates was motivated by prior HRI work showing that such characteristics can influence disclosure-related outcomes \cite{eyssel2017disclosure}.
\begin{table*}[t]
  \centering
  \caption{Summary of Final Stepwise Linear Regression Models for Post-Study Survey Scales.}
  \label{tab:survey_regression_summary}
  \footnotesize 
  \setlength{\tabcolsep}{4pt}
  \renewcommand{\arraystretch}{1.0}

  \begin{threeparttable}
  \begin{tabularx}{\textwidth}{l c c c X}
    \toprule
    \textbf{Scale} &
    \textbf{Model Fit} &
    \textbf{Condition $\beta$ ($p$)} &
    \textbf{Sig. Interaction $\beta$ ($p$)} &
    \textbf{Sig. Covariates $\beta$ ($p$)} \\
    \midrule

    \makecell[t]{\textbf{Disclosure} }  &
    \makecell[t]{Adj.\ $R^2=0.326$ \\ $F(9,40)=3.64$, $p=.002^{**}$} &
    \makecell[t]{\textbf{0.94} $(.010^{*})$ \\ (LD $>$ HD)} &
    \makecell[t]{\textbf{$-1.84$} $(.005^{**})$ \\ Cond.\ $\times$ roboTeamExp$_z$} &
    \makecell[tl]{gMale: $1.05$ $(.010^{**})$; \\ roboTeamExp$_z$: $1.28$ $(.001^{***})$ \\ roboExp$_z$: $-0.92$ $(.018^{*})$; \\ techExp$_z$: $-0.42$ $(.033^{*})$} \\

    \midrule
    \makecell[t]{\textbf{Teaming} }  &
    \makecell[t]{Adj.\ $R^2=0.168$ \\ $F(8,41)=2.24$, $p=.044^{*}$} &
    \makecell[t]{\textbf{0.93} $(.016^{*})$ \\ (LD $>$ HD)} &
    \makecell[t]{\textbf{$-1.24$} $(.068^{\bullet})$ \\ Cond.\ $\times$ roboTeamExp$_z$} &
    \makecell[tl]{roboTeamExp$_z$: $0.86$ $(.028^{*})$; \\ age$_z$: $0.44$ $(.037^{*})$ \\ roboExp$_z$: $-0.99$ $(.019^{*})$; \\ techExp$_z$: $0.53$ $(.058^{\bullet})$} \\

    \midrule
    \makecell[t]{\textbf{Rapport} \\ \textbf{(Coordination)}} &
    \makecell[t]{Adj.\ $R^2=0.201$ \\ $F(8,41)=2.54$, $p=.024^{*}$} &
    \makecell[t]{\textbf{0.86} $(.022^{*})$ \\ (LD $>$ HD)} &
    \makecell[t]{\textbf{$-0.96$} $(.022^{*})$ \\ Cond.\ $\times$ roboTeamExp$_z$} &
    \makecell[tl]{roboTeamExp$_z$: $0.68$ $(.005^{**})$; \\ age$_z$: $0.32$ $(.011^{*})$ \\ roboExp$_z$: $-0.64$ $(.010^{**})$} \\

    \midrule
    \makecell[t]{\textbf{Rapport} \\ \textbf{(Connection)}} &
    \makecell[t]{Adj.\ $R^2=0.127$ \\ $F(10,39)=1.71$, $p=.112$} &
    N/S &
    \makecell[t]{\textbf{$-0.79$} $(.030^{*})$ \\ Cond.\ $\times$ roboTeamExp$_z$} &
    N/A \\

    \midrule
    \makecell[t]{\textbf{Boredom} }  &
    \makecell[t]{Adj.\ $R^2=0.096$ \\ $F(11,38)=1.48$, $p=.181$} &
    N/S &
    \makecell[t]{\textbf{1.63} $(.042^{*})$ \\ Cond.\ $\times$ roboExp$_z$} &
    N/A \\

    \bottomrule
  \end{tabularx}

\begin{tablenotes}[para,flushleft]\footnotesize
\textbf{Sig.:} $^{***}p<.001$, $^{**}p<.01$, $^{*}p<.05$, $^{\bullet}p<.10$ (marg.). \textbf{N/A} = not applicable; \textbf{N/S} = not significant; $N=50$. 
\textbf{Abbrev.:} roboTeamExp$_z$ = prior experience teaming with robots; techExp$_z$ = prior experience with technology; roboExp$_z$ = prior experience with robots; gMale = gender (Male coded as 1); age$_z$ = age.
\textbf{Coding/Std.:} Covariates z-scored; Condition: HD ref. (LD--HD); gMale ref.=Female. 
\textbf{Selection/Diag.:} Stepwise AIC (both); VIF$<6$; normality, linearity, homoscedasticity OK.
\end{tablenotes}
  \end{threeparttable}
\end{table*}

%
\section{Results}
\label{sec:results}\subsection{Manipulation Check}
\label{section:results-check}
To confirm that the robot's disclosure behaviors differed between conditions, we analyzed 
the 
robot's total spoken word count, 
number of robot turns 
and 
self-disclosure levels 
for its utterances. 
A Student's t-test showed that the 
\textbf{robot word count} 
was significantly lower in condition LD
$(M=610.44, SD=206.42)$
than in condition HD
$(M=745.40, SD=241.22)$, 
$t(48)=-2.125,p=.039$.
A Wilcoxon rank-sum test revealed
the LD robot produced a significantly higher rate of 
\textbf{no disclosure utterances}
$(M=0.83, SD=0.14)$
compared to the HD robot
$(M=0.54, SD=0.14)$, 
$U=49.00, p < .001$.
%
Conversely, the HD robot generated 
significantly higher rates of 
\textbf{low-disclosure utterances} $(M_{LD} = 0.17, SD_{LD} = 0.13; 
M_{HD} = 0.39, SD_{HD} = 0.13)$, 
$U = 551.0, p < .001$, 
and 
\textbf{high-disclosure utterances} $(M_{LD} = 0.01, SD_{LD} = 0.02;
M_{HD} = 0.07, SD_{HD} = 0.10)$, 
as shown by Wilcoxon rank-sum test,
$U = 495.5, p < .001.$
Together, these results show that the disclosure manipulation was adequate:
the HD robot both spoke more and disclosed more. 
Additionally, there were no significant differences between conditions for the total
\textbf{number of robot turns} 
$(t(47)= 0.61, p= .54)$, 
\textbf{number of user turns}  
$(t(48)= 0.46, p= .65)$
and 
\textbf{user word count} $(t(48) = 0.31, p= .76)$.
Thus, differences in robot word count alone should not be interpreted as conversational dominance.
\label{sec:results}\subsection{No Condition Differences in Utterance Types}
To examine whether the robot's disclosure strategy influenced the distribution 
of utterance types, we compared the proportions of questions, imperatives, and 
comments (non-questions) produced by both users and the robot.
No significant differences were found between conditions for 
\textbf{User Utterance Types}: 
questions $(t(48) = 1.80, p = .08)$,
imperatives $(t(30) = 0.40, p = .70)$, 
and 
comments $(t(48) = -1.67, p = .10$).
No significant differences emerged for 
\textbf{Robot Utterance Types}: 
questions $(t(48) = -0.74, p = .47)$,
imperatives $(t(44) = -1.01, p = .32)$,
and 
comments $(t(48) = 1.31, p = .20)$.
These results indicate that the disclosure manipulation did not affect the 
structural composition 
of the conversation 
in terms of speech act distribution.

\subsection{Low Robot Disclosure Leads to High User Disclosure}
\label{section:b}
We explore how robot disclosure and robot conversational initiations (\eg prompted) influenced user disclosure. 
To examine whether robot disclosure predicted observed user disclosure, 
we fit a cumulative link mixed-effects model (logit link; Laplace approximation)
with
turn-level \textit{User Self-Disclosure Scores} (ordinal; see Sec. \ref{conversational-measures}) 
as the outcome. 
Fixed effects were
\textit{condition} (LD/HD)
and 
\textit{user (un)prompted responses},
with random intercepts for participant.
The model revealed that users who interacted with the HD robot disclosed less overall than those in condition LD ($\beta = -0.46, p= .042$). 
Turns where the user was prompted by the robot showed higher disclosure scores than unprompted turns ($\beta = 1.60, p < .001$).
The interaction between \textit{condition} and \textit{user prompted responses} was significant ($\beta = 0.45, p= .026$); 
this demonstrates that while participants in condition HD disclosed less overall, the positive effect of being prompted was stronger for them compared to those who interacted with the LD robot.

\subsection{Prior Teaming Experience Shaped Perceived Disclosure, Teaming and Rapport}
\label{section:results-quetionnaire}
We modeled each post‑study scale using 
linear regressions with 
Condition (LD/HD)
and 
demographic covariates (\eg age, gender, experience with tech) 
as predictors to examine how the robot's disclosure affected participants' subjective evaluations. 
Continuous covariates were z-scored prior to analysis; categorical covariates were entered as factors.
See Table \ref{tab:survey_regression_summary} for more details.
\begin{itemize}[leftmargin=*]
    \item \textbf{Disclosure Scale:} Users in LD reported higher disclosure scores than those HD ($\beta= 0.94$, $p=.010$). 
    A significant interaction with \textit{prior experience teaming with robots} indicated that more highly experienced participants reported less disclosure when interacting with the HD robot ($\beta= -1.84$, $p= .005$)\ 
    \item \textbf{Teaming Scale:} Users in LD reported more strongly on their perception of teaming ($\beta= 0.93$, $p= .016$). 
    \item \textbf{Rapport-Coordination Scale:} 
    Users in LD yielded higher coordination scores than in HD ($\beta= 0.86$, $p= .022$). 
    A significant interaction with \textit{prior experience teaming with robots} ($\beta= -0.96$, $p= .022$) suggested that prior experience moderated this effect, with more experienced participants showing reduced coordination benefits in the HD condition.
    \item \textbf{Rapport-Connection \& Boredom Scales:} 
    We observed no main effects of condition on these scales.
    However, significant interactions with prior experience emerged: 
    for Connection, an interaction with \textit{prior experience teaming with robots} ($\beta= -0.79$, $p= .030$) indicated that experienced participants reported lower connection in condition HD; 
    for Boredom, an interaction with \textit{prior robot experience} ($\beta= 1.63$, $p = .042$) suggested that experienced participants reported more boredom in the condition HD. 
\end{itemize}

\section{Discussion}
In this work, we examined how a non-anthropomorphic robot's disclosure strategy affects human disclosure in social communication during a cooperative task. 
We situated small talk concurrently with the physical task 
because our focus is social behavior integrated into ongoing work rather than separate social phases, reflecting realistic deployments where dedicated off-task interaction is limited, extending prior work on task-accompanying small talk \cite{pineda2025see}.
Although multiple factors shape disclosure,
the primary systematic difference between conditions was the robot's disclosure strategy, 
implemented through standardized prompts that guided the robot to share
more (HD) 
or
less (LD)
about itself.

We found that participants disclosed more in the 
\textit{LD} condition, despite the robot sharing more in the \textit{HD} condition.
When controlling for covariates, participants' perceived and labeled self-disclosure levels were significantly lower when interacting with the high-disclosure robot.
This occurred despite the robot in condition HD disclosing more frequently with increasing personalness and elaboration than the robot in condition LD; 
the robot had a higher word count and provided more low and high disclosure utterances.
Additionally, participants with more prior robot teaming experience were less receptive to high disclosure, 
and the low-disclosure robot yielded higher teaming and coordination scores.
To better understand the conversational dynamics underlying these patterns, we draw on post-study interviews with participants to enrich our discussion and interpretation of these findings.

These results contradict prior work, which has shown that increased self-disclosure from dialogue agents elicits reciprocal sharing \cite{ravichander2018empirical}.
Prior studies focusing on more human-like robots and virtual agents have shown that disclosure can successfully foster trust, empathy, and rapport \cite{chen2024effects, Tsumura_Yamada_2023}.
Below, we discuss three possible explanations for why our findings diverged from this established pattern: 
(1) embodiment incongruence between the robot's form and its disclosure strategy, 
(2) participants' perceptions of the robot's disclosures as inauthentic, and 
(3) privacy concerns triggered by the robot's sharing of past conversations.

\subsection{Embodiment Incongruence}

One possible explanation is that the robot's personal disclosures were incongruent with its non-humanoid form, 
creating a mismatch between its embodiment and conversational style. 
While a prior study found no effect of robot identity on self-disclosure in a real-world setting \cite{neerincx2022disclosure}, 
our findings with a non-humanoid manipulator suggest that embodiment may moderate how disclosure is received. 
This interpretation is supported by participants' lower teaming and coordination scores in the HD condition, suggesting that the disclosure mismatch degraded perceptions of effective collaboration.
However, it is worth noting that participants across both conditions valued 
the robot's conversational presence during the repetitive task. 
Some stated the robot \textit{``made the task more entertaining. Because, honestly, I was really 
bored" [L262],} suggesting that while disclosure strategy shapes teamwork 
perceptions, the presence of social interaction itself remains beneficial 
regardless of embodiment or disclosure level.

\subsection{Perceived Authenticity and Limits of Reciprocity}
One possible reason for why our results were the opposite of what was expected based on insights from the 
literature \cite{ravichander2018empirical, liang2024Dialoging}
is that participants may not have perceived the robot's self-disclosure as a genuine form of sharing. 
This perception may have arisen because the robot lacked human-like traits. 
Unlike human-like robots or agents, it might not have matched user expectations for a partner that discloses.

A review of post-study interviews revealed a mix of perceptions, with some participants seeing the robot's disclosures differently from others.
While participants did elaborate upon the robot's self-disclosure tendencies regardless of condition, there are signs that show people in condition HD did not perceive this robot to have engaged in true self-disclosure. 
One person hinted at wanting to know more details about the robot and its attributes, rather than its lived experiences, a key dimension of our self-disclosure manipulation.
They stated the robot mentioned how many degrees of freedom it had, but did not share much about itself otherwise [H230].
Another participant shared a similar sentiment, identifying a lack of disclosure from the robot,
\textit{``I spent a lot of time just asking it questions... I was curious if it would like talk about itself. It didn't seem to want to, or know how ... if I asked about itself, it felt like it would kind of deflect" [H204].}
In contrast, one participant who interacted with the low-disclosure robot stated, \textit{``when I started talking to the robot more... it started saying things about itself, which I found was really funny" [L223].}
Therefore, despite the robot talking about its supposed past conversations with other people, participants perhaps may not have equated this as a form of self-disclosure.

Furthermore, one user indicated how they wanted the robot to share first before they would disclose, implying that they did not feel the high-disclosure robot was adequately disclosing, 
\textit{``it's just like trying to trick me into talking. 
So before expecting me to share, I would expect it to share" [H242].} 
This notion aligns with prior work on individuals' tendencies to disclose, even with strangers, if they are the recipients of disclosure first \cite{nass2000machines}.
Thus, our intended manipulation might not have been as strong as we intended.

Although participants disclosed more when directly prompted to respond by the robot, 
our findings suggest that this reciprocity was narrow in scope and did not generalize to broader engagement across the interaction.
Some participants described the robot's responses as short, often only reaching one or two sentences in length, which impacted the flow of conversation.
One participant who interacted with the high-disclosure robot noted, 
\textit{``It would tell me things that other people had told it, but it would only give very generic answers... if I asked more, it seemed to run out of details, like it wouldn't remember'' [H204]}. 
Another participant reflected on this dynamic as a kind of ``conversational ping pong,'' explaining that the robot could not carry topics forward in the conversation,
\textit{``It was like playing conversational ping pong... if I brought up a subject, it would just throw it back at me'' [H204]}. 
Others described the exchanges as more mechanical than social, with one participant from condition HD stating,
\textit{``It did not feel like a human conversation. It felt like a questionnaire'' [H265]}. 

These reflections highlight the limits of reciprocity in our study. 
While the robot's prompts successfully elicited disclosure in the moment, participants questioned the authenticity and depth of the conversational exchange. 
One participant described ignoring the robot's questions about hobbies and favorite color because 
\textit{``it just didn't feel genuine.''} 
They explained on their own,
without prompting from the interviewer, 
that personal disclosure usually works through reciprocity, 
\textit{``we tell them what our favorite color is, and we hope that they'll tell us theirs. I know it doesn't have a favorite color... hobbies... so it just felt very disingenuous---kind of pointless for me to respond'' [L241].}  
As this participant noted, 
without genuine reciprocity, the robot's questions felt superficial, 
underscoring that reciprocity in our study was limited to prompted exchanges and failed to support broader engagement.
Consistent with this, participants rated the HD robot 
as a less effective teammate, with lower coordination scores despite the 
robot's attempts at personal sharing.

\subsection{Privacy Concerns with Self-Disclosure}

Another possibility for why participant self-disclosure decreased in the HD condition relative to the LD condition could be that this effect may have been directly linked to our disclosure manipulation itself: 
the robot's sharing of past conversations. 
In post-study interviews,
users from both conditions acknowledged the potential privacy concerns of disclosing to a conversational robot due to the creation of a digital footprint.
One participant who interacted with the low-disclosure robot stated, 
\textit{``As an employee that's doing these things [in an actual workplace], a concern that I would have would be like my data and its privacy. 
You know, chatbots are listening all the time, and they're obviously processing what I'm saying... I have no idea where any of it is going. 
Like, what if I like, trash talk my boss, except it's accessible to my boss'' [L241].}
While this real concern is non-unique to any one of our experimental conditions, a potential implication of data and privacy leakage was more apparent in the HD condition because the robot shared details of (fictional) prior conversations with every participant. 
One person who interacted with the high-disclosure robot explained, 
\textit{``I didn't like that aspect because I felt that like it's just gonna share what I say to like another worker later on. So I didn't say anything afterwards" [H242].}

Research has highlighted that attitudes towards privacy can serve as a factor for disclosure and realizing its benefits among social robot interactions \cite{dietrich2023disclose}.
Participants have also expressed concerns with sharing personal information in recent experiments with LLM-based social robots \cite{Irfan_Skantze_2025}. 
Thus, our intentional design decision to 
enable the robot to share about its (fictional) past conversations may have directly backfired, prompting less self-disclosure from participants.

\subsection{Other Factors Shaping Self-Disclosure}

Our quantitative analyses revealed two additional categories of factors that influenced user self-disclosure patterns: individual differences and prompted responses.

\subsubsection{Prior Robot Teaming Experience}
Our analyses suggest that participants with more prior robot teaming experience were less receptive to the high-disclosure robot. 
This may reflect that experienced users carry stronger mental models of robots, making them less inclined to accept persona-based disclosure as authentic. 
For example, one participant noted that they viewed the robot's conversational and task-oriented behaviors as 
\textit{``completely two different tasks'' [H231]}. 
These findings highlight that user background, 
particularly prior experience with robots,
can shape how disclosure strategies are received.

\subsubsection{Prompted vs Unprompted Responses}
The way disclosures were initiated also shaped interaction patterns. 
Participants in the high-disclosure condition rarely disclosed spontaneously, but they did share when directly prompted by the robot.
Prompting thus narrowed the gap between conditions, 
eliciting disclosure in the moment without increasing overall conversational engagement.
This aligns with prior communication research showing that disclosure often clusters around specific conversational turns rather than being distributed evenly \cite{pearce1973self}. 
\subsection{Design Implications}
Our results translate into three design implications for integrating self-disclosure into physical, non-anthropomorphic human–robot teaming: 
\designimp{Design Implication \#1 (Match Disclosure to Form)}{Reserve richer self-disclosure for human-like robots; on non-anthropomorphic manipulators, favor light, functional, contextually grounded small talk.}
\designimp{Design Implication \#2 (Elicit, Don't Over-Disclose)}{Use targeted prompts to draw out user disclosure rather than extensive persona-based robot sharing, which can read as inauthentic on functional platforms.}
\designimp{Design Implication \#3 (Account for User Experience)}{Adapt robot social behavior; users experienced in robot teaming are less receptive to persona-based disclosure.}

\section{Limitations}
This study used a simulated quality control task with a participant pool drawn largely from university students and employees. 
Findings may therefore not generalize to broader populations or real industrial contexts, where time pressure, task severity and privacy concerns may further alter disclosure dynamics; 
participants' prior robot experience may have also impacted generalization. 
Our disclosure manipulation relied on a fixed set of predefined details that the robot could draw from when triggered to respond.
While the robot decided autonomously what to disclose, the instructions within its LLM prompt may have limited the authenticity and flexibility of its self-disclosures compared to more adaptive methods.
Finally, we studied a single non-humanoid manipulator robot.
Accordingly, our interpretation of non-anthropomorphism is
specific to the platform tested here and should not be generalized
to all non-anthropomorphic robots, which may differ in morphology,
motion style, role expectations, and social framing. 
Future work
should explore disclosure across different robot forms, varying
social features,
the timing of conversation relative to the task (\eg concurrent vs. before/after or during breaks),
and in longitudinal settings.
\section{Conclusion}
Our work explored the intricacies of integrating small talk into interactions with a non-humanoid robot in a cooperative setting. 
Disclosure from the robot did not always encourage reciprocity, showing that disclosure is a double-edged strategy; our analysis revealed 
lower user self-disclosure with the high-disclosure robot. 
By combining quantitative models with qualitative interviews, this work contributes empirical evidence that disclosure must be carefully calibrated to context and user expectations, especially in tasks with non-anthropomorphic robots. 
This work advances the understanding of how to design disclosure strategies that support, rather than hinder, human-robot teamwork and cooperation.

\section*{Acknowledgments}

\textbf{Author CRediT.}
Concept. \& Method. (KTP, ZX, CH); Software (KTP, BC); Validation (KTP); Formal Analysis (KTP, AM, AG, TW); Investigation (KTP, AM, BC); Data Curation (KTP, AM, BC, AG, TW); Original Draft (KTP, AM, BC); Review \& Editing (all); Vis. (KTP, AG); Supervision (ZX, CH); Project Admin. (KTP); Funding (CH).

\textbf{Funding Source:} National Science Foundation award \#2141335.  \textbf{AI Statement:} Text edited with LLM; output checked for correctness by authors. 

\balance

\bibliographystyle{IEEEtran}
\bibliography{IEEEabrv,references}

\end{document}